\documentclass[letterpaper,10pt,conference]{ieeeconf}
\IEEEoverridecommandlockouts
\usepackage[table,dvipsnames]{xcolor}
\usepackage{graphicx}
\usepackage{booktabs}
\usepackage{float}
\usepackage{multicol}
\usepackage{subcaption}
\usepackage{authblk}
\usepackage{multirow}
\usepackage{todonotes}

\usepackage[T1]{fontenc}
\usepackage{caption}
\usepackage[table]{xcolor}

\makeatletter
\let\NAT@parse\undefined
\makeatother
\usepackage{cite}
\usepackage{hyperref}
\hypersetup{
  hypertex=true,
  colorlinks=true,
  linkcolor=blue,
  anchorcolor=blue,
  citecolor=blue
}

\title{\LARGE \bf
CoinRobot: Generalized End-to-end Robotic Learning for Physical Intelligence
}

\author{
  Yu Zhao$^{1}$, Huxian Liu$^{1}$, Xiang Chen$^{2}$, Jiankai Sun$^{3}$, Jiahuan Yan$^{1}$, Luhui Hu$^{1}$ \\
  $^{1}$ZhiCheng AI, $^{2}$Peking University,  $^{3}$Stanford University
}

\begin{document}
\maketitle
\thispagestyle{empty}
\pagestyle{empty}



\begin{abstract}
Physical intelligence holds immense promise for advancing embodied intelligence, enabling robots to acquire complex behaviors from demonstrations. However, achieving generalization and transfer across diverse robotic platforms and environments requires careful design of model architectures, training strategies, and data diversity. Meanwhile existing systems often struggle with scalability, adaptability to heterogeneous hardware, and objective evaluation in real-world settings. We present a generalized end-to-end robotic learning framework designed to bridge this gap. Our framework introduces a unified architecture that supports cross-platform adaptability, enabling seamless deployment across industrial-grade robots, collaborative arms, and novel embodiments without task-specific modifications. By integrating multi-task learning with streamlined network designs, it achieves more robust performance than conventional approaches, while maintaining compatibility with varying sensor configurations and action spaces. We validate our framework through extensive experiments on seven manipulation tasks. Notably, Diffusion-based models trained in our framework demonstrated superior performance and generalizability compared to the LeRobot framework, achieving performance improvements across diverse robotic platforms and environmental conditions. 

\end{abstract}

\section{Introduction}
Recent studies \cite{c1,c2,c45} have shifted their attention toward exploring the applications of imitation-based techniques in the field of robotic control and manipulation. This trend is largely influenced by the expanding role of generative artificial intelligence across various industrial sectors.\cite{c48} \cite{c49}


To evaluate the effectiveness of the proposed framework, we designed 7 distinct robotic tasks, each characterized by specific features tailored to real-world conditions. Section \ref{sec:Task Design} delves into the requirements and methodologies involved in the design of these tasks, while Section \ref{sec:Experiments and Results} systematically examines how task-specific characteristics impact performance outcomes in real-world testing scenarios.

Many researchers are exhausted by the repetitive and time-consuming task of adapting to new robots or modifying robot systems for different model training platforms. Given this challenge, it is highly meaningful to construct a general platform that can seamlessly integrate new types of robots and switch to new model training platforms as needed.
Thus, we propose CoinRobot as a solution to these issues. In this paper, we will demonstrate the architecture of CoinRobot and how it achieves decoupling from specific robot types and model training platforms. Additionally, we will highlight the ease with which CoinRobot can incorporate new robot types and adapt to new model platforms.

In addition to the versatility of using a general-purpose robotic arm that can meet the demands of various industrial scenarios, we also successfully enable a single checkpoint to perform multiple tasks by combining datasets and applying minor adjustments to the training strategy, which are discussed in Section \ref{sec:Model Generalization}.
\begin{figure}
    \centering
    \includegraphics[width=\linewidth]{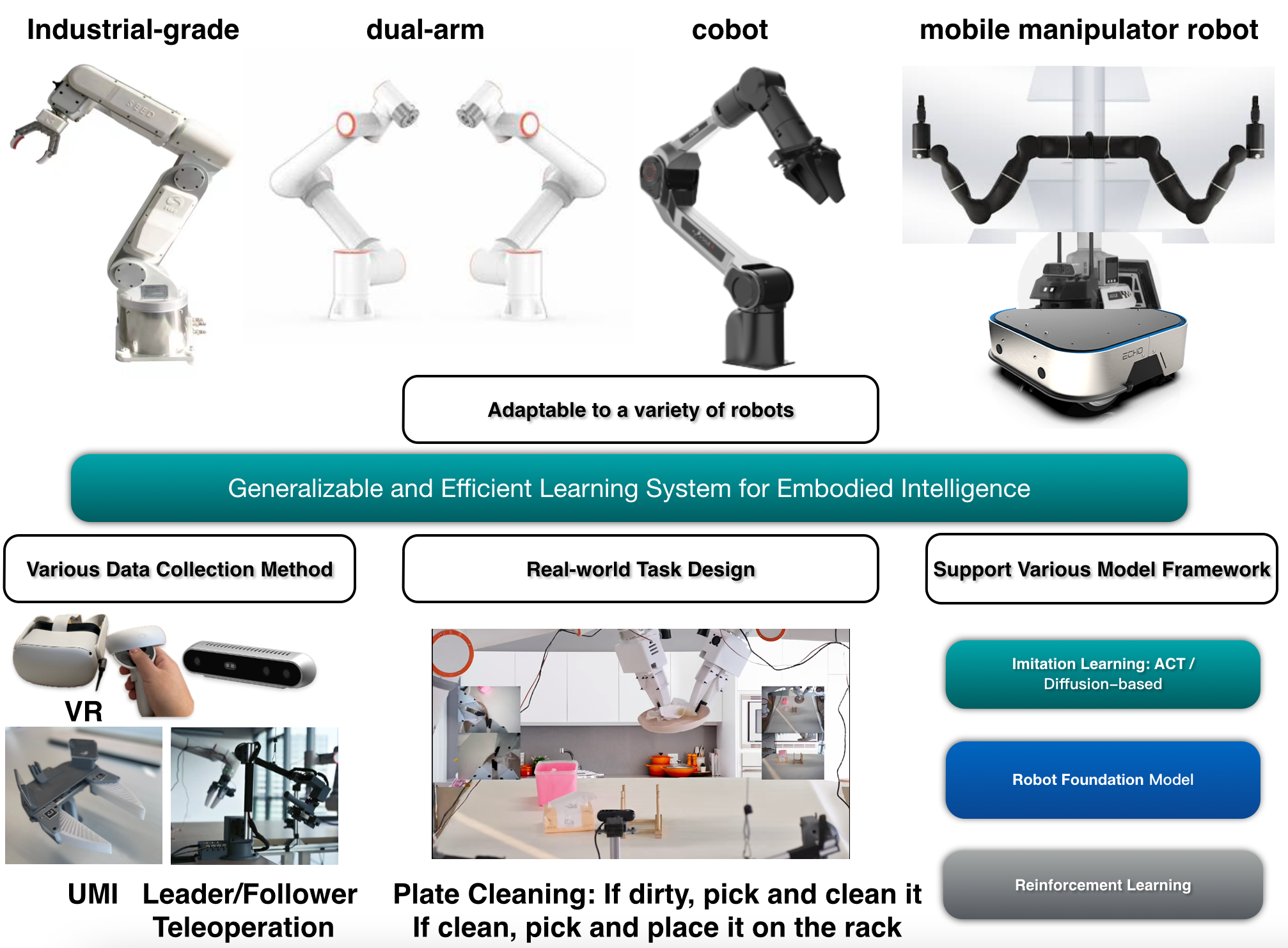}
    \caption{Overview of the framework: A real-world robot learning setup can be constructed using a variety of robots, data collection method and adaptable to multiple model structure}
    \label{fig:enter-label}
\end{figure}
In addition to our findings, we are releasing the complete dataset and trained models using our framework to support researchers and practitioners interested in utilizing the same robotic interface. While we provide these resources to facilitate replication and further exploration, we strongly encourage others to develop their own instances. Our work demonstrates that with various robotic systems, it is possible to achieve robust and effective results.
The main contributions can be summarized as:
\begin{itemize}
    \item We propose CoinRobot, a platform designed for data collection and inference in general robotics. CoinRobot supports the rapid and low-cost addition of new robot types and sensors, while significantly reducing the burden of adapting data collection and inference processes for new robots or switching to new model platforms.
    \item We designed and collected over 3,000 episodes across 7 distinct real-world robotic tasks, to verify our findings on the correlation between task difficulty and performance.
    \item We demonstrate model generalization by successfully enabling task adaptation through minimal dataset integration and slight modifications to the training process.
    \item We share complete setup details, training insights, and lessons learned to support reproducibility.
\end{itemize}

\begin{figure*}[ht]
    \centering
    \includegraphics[width=\textwidth]{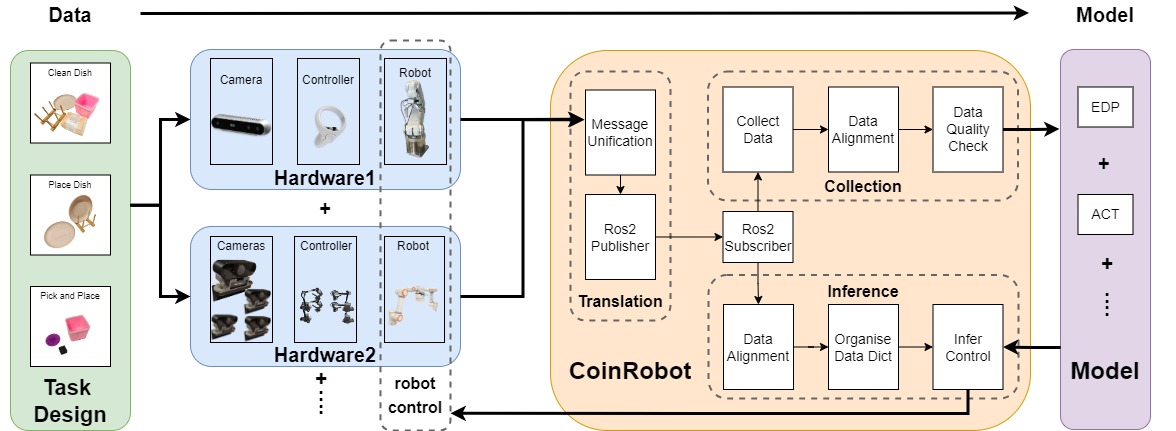}
    \caption{End-to-End Framework: The pipeline illustrates the end-to-end process for a generalized robotic learning implementation, from hardware setup and task design to data collection, modeling and training, evaluation , and model deployment. This framework is designed to be structurally simple and economically feasible for deployment.}
    \label{fig:framework}
\end{figure*}

\section{Related Work}
\subsection{End-to-End Robot Learning} 
A central challenge in robotic learning is the limited availability of diverse and extensive training datasets. End-to-end robot learning has emerged as a promising solution, offering a streamlined and data-driven approach that leverages imitation learning to simplify the training process \cite{c54}. While projects such as ALOHA \cite{c13} and UMI \cite{c36} have demonstrated success in learning from human demonstrations, they often require specialized, professional-grade setups or expensive robotic hardware, limiting their accessibility. Similarly, frameworks like LeRobot \cite{cadene2024lerobot} provide modular training infrastructure but do not prioritize adaptability across diverse hardware or real-world deployment scenarios.

In this work, we introduce an end-to-end robot learning framework designed to address these limitations. Our approach emphasizes rapid adaptability to a wide range of robotic hardware, training model architectures, and data collection methodologies. By prioritizing flexibility and ease of use, our framework enables seamless integration with various tools and environments, making it accessible to non-experts while maintaining robustness under suboptimal conditions. This adaptability ensures that our solution can be readily applied across diverse use cases, lowering the barrier to entry for beginners and expanding the practical reach of robot learning.

\subsection{Imitation Learning} 
Imitation Learning (IL) \cite{c5,c6} is a prominent approach in robotics and autonomous systems, enabling agents to acquire complex behaviors by mimicking expert demonstrations \cite{c7,c8,c9}. Among the various IL techniques, behavioral cloning (BC) has been widely used, framing the task as a supervised learning problem where actions are directly mapped from perceptions \cite{c25,c26}. A well-known example is the ACT policy \cite{c13}, which exemplifies explicit policy learning. However, explicit policies often struggle with multimodal behavior \cite{c24}, as they tend to perform poorly in scenarios with diverse demonstrated actions. 
Diffusion-based policies \cite{c1, c51, davies2024spatiallyvisualperceptionendtoend, huang2025robograspuniversalgraspingpolicy} addresses the instability of implicit policies by modeling the gradient field of an implicit action score, eliminating the need for negative sampling and improving training stability. We have tried both Denoising Diffusion Probabilistic Models (DDPM) \cite{c52} Denoising Diffusion Implicit Models (DDIM) \cite{c55} as paradigms for action prediction, and opt for DDIM as it decouples the number of denoising iterations in training and inference, thereby allowing the algorithm to use fewer iterations for inference to speed up the process, see Section \ref{sec:Robot Policy System}. 

\section{Framework Setup}

\subsection{Hardware Preparation}
Our hardware devices for data collection and model deployment facilities are listed as follows:
\begin{itemize}


    \item \textbf{Robotic Arm:} The dual robotic arm used in our experiments is from FAIR INNOVATION(<\$5,600) and Machine type of this robot is FR5.The FAO FR5 robotic arm is a 6-DOF collaborative robot with a 5 kg payload capacity and a reach of 922 mm. The single robotic arm used in our experiments is Seed Robot, an industrial-grade robot arm with moderate control frequency, localization precision, and motion planning capabilities.
    \item \textbf{Cameras:} For dual-arm tasks, four Hikvision web RGB cameras were used to capture visual information. Two cameras were mounted on the end-effectors of the dual robotic arms, respectively, to provide close-up views. The  other two camera was positioned at a high and a lower position to offer a global view and depth information for the task. All cameras captured images at a resolution of 640×320 at a frequency of 30 Hz. CoinRobot collected all these data, but we may downsample or resize the images according to the model's requirements. For single-arm tasks, two Intel RealSense D415 RGB-D cameras were utilized for frame acquisition. One camera was mounted on the end-effector of the robotic arm to provide a close-up perspective, while the second was positioned to offer a global view of the workspace.
    \item \textbf{Controller:}
    For dual-arm tasks we assembled our in-house designed leader bot, whose joint-to-joint length ratio matches that of the follower bot. This design enables intuitive control of the follower bot through joint angles. To capture the joint angles of the leader bot for control purposes, we installed Fashion Star HP8-U45H-M motors at each joint. The joint angles of the leader bot are recorded at a frequency of 160 Hz. For single-arm tasks an Oculus Quest 2 headset, along with its controller, was used for data collection. 
\end{itemize}


We summarize the hardware we have used and supported by the time we release this work in Table \ref{tab:Comparison}.

\begin{table}[]
\begin{tabular}{cccc}
\hline
\textbf{Robot Name}                                             & \textbf{DOF} & \textbf{Category}                                               & \textbf{Price (USD)} \\ \hline
FAIRINO FR5                                                    & 6            & Industrial-grade  & 5,600(dual-arm)               \\
\begin{tabular}[c]{@{}c@{}}AgileX Piper \\ \end{tabular} & 6 & Cobot  & 2,000    \\Realman Gen72                                                       & 7   & Cobot                   & 1,400               \\
Seed Robot                                     & 6            & Industrial-grade                                                       & 2,200       \\ \hline
\textbf{Data Collection Tool}                                                  & \multicolumn{2}{c}{\textbf{Quantity}}                                                      & \textbf{Unit Price(USD)}      \\ \hline
\begin{tabular}[c]{@{}c@{}}Intel RealSense \\ D415\end{tabular} & \multicolumn{2}{c}{2}                                                           & 209.99               \\
Oculus Quest 2                                                  & \multicolumn{2}{c}{1}                                                           & 299.99               \\
\begin{tabular}[c]{@{}c@{}}Hikvision RGB cameras \\ DS-E12a\end{tabular}
            & \multicolumn{2}{c}{4}                                                           & 20               \\
Leader Arm(Proprietary)                                                  & \multicolumn{2}{c}{2}                                                           & -               \\\hline
\end{tabular}
\caption{Our framework supports a variety of robots and data collection method}
\label{tab:Comparison}
\end{table}

\subsection{Software System}
Code Adaptation for data collection and inference can often be the most tedious part of the research process. When dealing with a new type of robot, researchers must undertake numerous trivial tasks to gather observations for collection and inference. To decouple the robots, sensors and model platform from the data collection and inference, we introduce CoinRobot, a comprehensive platform designed for data collection and inference tailored for general robotics applications. The CoinRobot platform is structured into three key components: Collection, Inference, and Translation.

The CoinRobot data collection module gathers data through ROS2 topic messages during the remote control operations based on ros2 between a leader bot and a follower bot. Developing a remote control system based on ROS2 offers several advantages; it notably reduces the development costs compared to other communication forms, as ROS2 adeptly manages issues related to multi-process communication and coordination. It is common for robot manufacturers to provide ROS2 packages for controlling their robots, which notably lowers the code development cost for researchers building their own robot remote control systems. Additionally, ROS2 facilitates seamless compatibility between different leader bots and follower bots, enabling the swift construction of new remote control systems whenever required.

The CoinRobot data collection module simplifies the process of adding new robot types by eliminating the need to adapt to different types of collected ROS2 messages. This is achieved through a translation module developed within CoinRobot, which subscribes to all relevant ROS2 topics, converts diverse ROS2 messages into a unified format, and reorganizes the data into dictionaries. These dictionaries are then converted into JSON messages and published through ROS2 topics. When integrating a new robot type, the collector only needs to develop a ROS2 package based on the provided code templates. This package, serving as the translation module in CoinRobot, converts the required ROS2 messages into the unified format and publishes them, ensuring seamless integration.

We timestamp each frame from all sensors and capture their data comprehensively, without initially addressing time alignment among sensors during data collection. After collecting the data, we verify that the average and standard deviation of the sampling frequency meet the predefined limits. Subsequently, we align all the data according to the recorded timestamps, following the self-designed alignment strategy.

The CoinRobot inference module reuses the same data-fetching module employed during data collection, significantly reducing the coding burden when integrating new sensor configurations. Researchers can easily obtain inferred actions from the model and instruct the robot to perform tasks according to their needs. Additionally, we have developed a model bridge template for the CoinRobot inference module, allowing seamless connection to any model platform.

\subsection{Data Collection}

Data collection involves using the master bot's handler to control the follower bot as it performs tasks. The master bot's intuitive design allows even untrained individuals to quickly learn how to operate it effectively, thanks to its precise and timely control over the follower bot. The follower bot publishes its state information at 60 Hz, while the master bot issues control instructions at 160 Hz. Additionally, cameras capture images at 30 Hz. During data collection, CoinRobot captures all sensor data at their respective frequencies. 
After collection, the data is aligned to a frequency no higher than the lowest frequency among all sensors. In practice, we set the aligned frequency at 10 Hz.

During data collection, one operator is involved. This operator is responsible for both operating the master bot to complete tasks and arranging objects based on specific scenarios and their own intuition. The collected data is tagged with the operator's name to distinguish between identical tasks performed by different collectors, a distinction that serves an important purpose.

\subsection{Robot Policy System}
\label{sec:Robot Policy System}


Our framework is designed to support a diverse range of model architectures, including those tailored for imitation learning, such as Action Chunking Transformers (ACT)\cite{c13} and diffusion-based models\cite{c1}. While ACT demonstrates proficiency in generating smooth action trajectories, we observed that it exhibits limitations in generalizing across multimodal distributions, which can hinder its effectiveness in complex scenarios.

We decompose the robot control policy into two distinct yet interconnected components: the perception module and the action prediction module. The perception module is responsible for processing sensory inputs from the physical environment and generating meaningful embeddings, while the action prediction module utilizes these embeddings to produce actionable trajectories for robotic control.

The perception module integrates multimodal data, including visual inputs from camera sources: for single robot arm, a wrist-mounted camera providing a first-person perspective and a fixed external camera offering a third-person view, for dual-arm robot, two wrist-mounted cameras and two (high and low) fixed external cameras offering third-person views. Additionally, low-dimensional state information such as end-effector pose is incorporated to enrich the robot's understanding of its environment. This combined data is processed by a deep learning network to generate compact embeddings. We evaluated several network architectures, and present the result in Table \ref{tab:task_performance}. Among these, the FPN-based ResNet34 \cite{c40, c41} demonstrated superior performance, effectively capturing multi-resolution visual features and significantly enhancing the quality of the generated embeddings.

The action prediction module is designed with mapping the encoded perception data into executable trajectories for robotic manipulation. In our implementation, we employ a Denoising Diffusion Implicit Models (DDIM) \cite{c52} as the core architecture. The DDIM decouples the number of denoising iterations in training and inference, thereby allowing the algorithm to use fewer iterations for inference to speed up the process. Within this framework, we explored two primary network architectures: Convolutional Neural Networks (CNNs) and Transformers, each offering unique advantages in terms of scalability and feature representation.

Together, these modules form a cohesive policy control system within our framework. Importantly, the architecture of each module is highly flexible, allowing for customization based on specific task requirements without being constrained to a predefined design. This adaptability ensures that our framework can accommodate a wide range of robotic applications while maintaining robust performance.

\subsection{Task Design}
\label{sec:Task Design}
\begin{figure*}
    \centering
    \includegraphics[width=\linewidth]{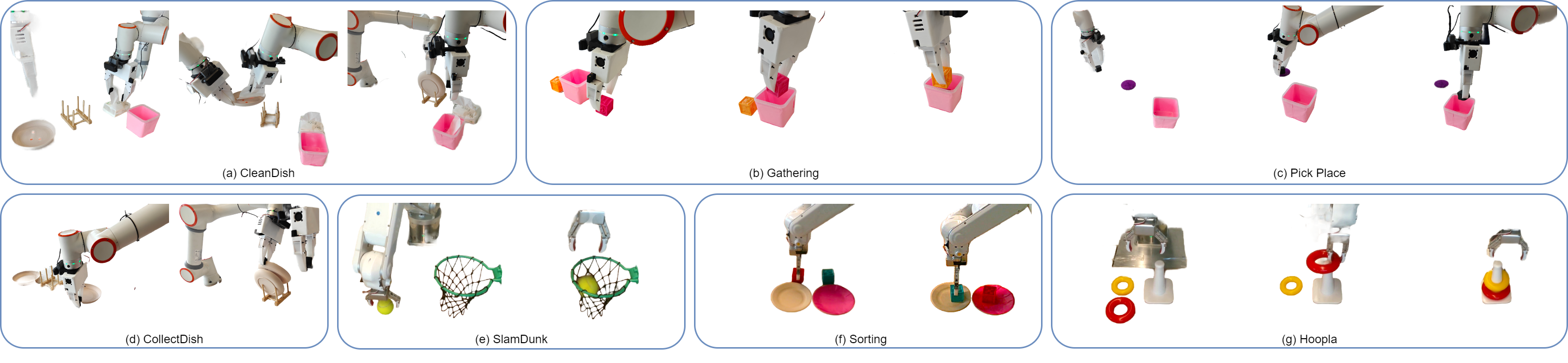}
    \caption{Real-world task design: Each task features distinct visual state changes. The left image of each subfigure shows the initial state of the
environment; the right image shows the goal state. See Section \ref{sec:Task Design} for a detailed task description.
    }
    \label{fig:task_workflow}
\end{figure*}

In this study, we present 7 well-defined real-world tasks, as shown in Figure \ref{fig:task_workflow}, using easily accessible objects to facilitate easy replication. Each task targets specific model capabilities. The following 4 tasks use dual-arm robot (FAIRINO FR5).
\textbf{CleanDish}: If the dish is dirty, the left arm picks up a tissue, while the right arm picks up the dish. The left arm, holding the tissue, then wipes the dish. After cleaning, the left arm places the dish on the rack. If the dish is already clean, it is placed directly on the dish rack without cleaning.
\textbf{Gathering}: The right arm grasps a cube and places it in a box. Subsequently, the left arm grasps another cube and places it in the same box.
\textbf{PickPlace}: The left hand grasps a cube and places it on a plate. The right hand then grasps the cube from the plate and places it in a box.
\textbf{CollectDish}: The right arm collects a dish and places it on a rack. The left arm then collects another dish and places it on the rack. 
The following 3 tasks use single-arm Seed Robot.
\textbf{SlamDunk}: The robot grasps a tennis ball and tosses it into a toy basket. 
\textbf{Sorting}: The robot picks up colored blocks and places them on plates of the same color.  \textbf{Hoopla}: The robot picks up a larger hoopla and places it on a stacking tower, followed by picking up a smaller hoopla and placing it on the same stacking tower. 
To introduce variability and prevent overfitting to specific configurations, each task incorporates a degree of randomness, such as variations in the initial positions of objects and receptacles. These unmeasurable variations encourage generalization and improve the model's robustness.

\subsection{Model Evaluation Metrics}
\label{sec:Evaluation Metric}
The assessment of autonomous policy performance adheres to a methodology consistent with established approaches outlined in prior research \cite{c2}. As highlighted in the literature, the loss curve observed during the training phase frequently exhibits a weak correlation with the policy's practical task performance. Specifically, the loss function employed in these studies typically quantifies the prediction error of noise within the diffusion process. However, we identify a critical limitation: this metric fails to provide a direct measure of the model's alignment with the desired action distribution. To address this, we propose the adoption of Mean Squared Error (MSE) in action prediction as a more informative evaluation metric. Additionally, we incorporate the visualization of action trajectories to facilitate an intuitive and immediate assessment of the model's training efficacy.

\section{Experiments and Results}
\label{sec:Experiments and Results}


\subsection{Task Analysis}

We conducted a comprehensive analysis to explore the relationship between task success rates and task-specific characteristics. To ensure that the observed outcomes were not influenced by variations in model architecture or training strategies, we maintained a consistent architecture throughout the experiments and selected the optimal model checkpoint based exclusively on performance metrics, independent of training duration.

Table \ref{tab:task_performance} provides a summary of manually extracted features for each task, alongside the corresponding dataset sizes. Our analysis reveals noteworthy insight 
 below:

\begin{enumerate}
\item \textbf{Action Types of Task}: Success rates demonstrate considerable variability across different tasks. However, tasks that seems quite different at first glance but actually share essentially the same action characteristics—such as pick-and-place operations—exhibit a notable property: models trained on one task can be efficiently fine-tuned to perform another related task with minimal additional training, often requiring only a few epochs, as noted in \ref{sec:Model Generalization}.  This observation suggests the presence of shared latent features within the action space of robotic tasks, indicating that the underlying representations of similar actions are inherently transferable.

\item \textbf{Visual Distinguishability}: Empirical observations indicate that tasks emphasizing color differentiation tend to achieve higher success rates when utilizing a ResNet-based perception encoder. 
\end{enumerate}

These findings provide valuable guidance for future research, highlight the potential for leveraging pretrained models and transfer learning to accelerate adaptation to new but related tasks. And it also emphasizes the need to tailor model architectures and training approaches to the specific characteristics of the task at hand. By leveraging these insights, we can better optimize robotic systems for improved performance across a diverse range of applications.



    

\subsection{Training Platform and Model Architecture Ablation Study}

\begin{table*}[ht]
\centering
\begin{tabular}{lcccccrrr}
\hline
\multicolumn{1}{c}{\multirow{2}{*}{\textbf{\begin{tabular}[c]{@{}c@{}}Task\\ Name\end{tabular}}}} &
  \multirow{2}{*}{\textbf{\begin{tabular}[c]{@{}c@{}}Object\\ Num\end{tabular}}} &
  \multirow{2}{*}{\textbf{Color}} &
  \multirow{2}{*}{\textbf{Logic Step}} &
  \multirow{2}{*}{\textbf{\begin{tabular}[c]{@{}c@{}}Avg\\ Length\end{tabular}}} &
  \multirow{2}{*}{\textbf{\begin{tabular}[c]{@{}c@{}}Demo\\ Num\end{tabular}}} &
  \multicolumn{3}{c}{\textbf{\begin{tabular}[c]{@{}c@{}}Success Rate\end{tabular}}} \\ \cline{7-9} 
\multicolumn{1}{c}{} &
   &
   &
   &
   &
   &
  \scriptsize\textbf{\begin{tabular}[c]{@{}l@{}}ACT\\ /\\ LEROBOT\end{tabular}} &
  \scriptsize\textbf{\begin{tabular}[c]{@{}l@{}}DP\\ /\\ LEROBOT\end{tabular}} &
  \scriptsize\textbf{\begin{tabular}[c]{@{}l@{}}DP\\ /\\ COINROBOT\end{tabular}}\\ \hline
PickPlace &
  3 &
  No &
  1 &
  13.32s &
  50 &
  30\% &
  0\% &
  \textbf{64\%} \\
\cline{1-9}
\end{tabular}

\centering
\begin{tabular}{lcccccrrrr}
\hline
\multicolumn{1}{c}{\multirow{2}{*}{\textbf{\begin{tabular}[c]{@{}c@{}}Task\\ Name\end{tabular}}}} &
  \multirow{2}{*}{\textbf{\begin{tabular}[c]{@{}c@{}}Object\\ Num\end{tabular}}} &
  \multirow{2}{*}{\textbf{Color}} &
  \multirow{2}{*}{\textbf{Logic Step}} &
  \multirow{2}{*}{\textbf{\begin{tabular}[c]{@{}c@{}}Avg\\ Length\end{tabular}}} &
  \multirow{2}{*}{\textbf{\begin{tabular}[c]{@{}c@{}}Demo\\ Num\end{tabular}}} &
  \multicolumn{4}{c}{\textbf{\begin{tabular}[c]{@{}c@{}}Success Rate\end{tabular}}} \\ \cline{7-10} 
\multicolumn{1}{c}{} &
   &
   &
   &
   &
   &
  \scriptsize\textbf{\begin{tabular}[c]{@{}l@{}}Res18\\ /\\ UNet\end{tabular}} &
  \scriptsize\textbf{\begin{tabular}[c]{@{}l@{}}Res18\\ /\\ TF\end{tabular}} &
  \scriptsize\textbf{\begin{tabular}[c]{@{}l@{}}Res34\\ /\\ TF\end{tabular}} &
  \scriptsize\textbf{\begin{tabular}[c]{@{}l@{}}FPN\\ /\\ TF\end{tabular}} \\ 
  \hline
SlamDunk &
  1 &
  No &
  1 &
  22.07s &
  100 &
  92\% &
  \textbf{92\%} &
  80\% &
  86.7\% \\
Sorting &
  2 &
  Yes &
  2 &
  24.02s &
  330 &
  63\% &
  \textbf{74\%} &
  56\% &
  70\% \\

Hoopla &
  2 &
  Yes &
  2 &
  40.28s &
  100 &
  26\% &
  60\% &
  70\% &
  \textbf{72\%} \\
\cline{1-10}
\end{tabular}

\centering
\begin{tabular}{lcccccrr}
\hline
\multicolumn{1}{c}{\multirow{2}{*}{\textbf{\begin{tabular}[c]{@{}c@{}}Task\\ Name\end{tabular}}}} &
  \multirow{2}{*}{\textbf{\begin{tabular}[c]{@{}c@{}}Object\\ Num\end{tabular}}} &
  \multirow{2}{*}{\textbf{Color}} &
  \multirow{2}{*}{\textbf{Logic Step}} &
  \multirow{2}{*}{\textbf{\begin{tabular}[c]{@{}c@{}}Avg\\ Length\end{tabular}}} &
  \multirow{2}{*}{\textbf{\begin{tabular}[c]{@{}c@{}}Demo\\ Num\end{tabular}}} &
  \multicolumn{2}{c}{\textbf{\begin{tabular}[c]{@{}c@{}}Success Rate\end{tabular}}} \\ \cline{7-8} 
\multicolumn{1}{c}{} &
   &
   &
   &
   &
   &
  \scriptsize\textbf{\begin{tabular}[c]{@{}l@{}}ACT(fixed position)\\ /\\ LEROBOT\end{tabular}} &
  \scriptsize\textbf{\begin{tabular}[c]{@{}l@{}}ACT(varied position)\\ /\\ LEROBOT\end{tabular}}\\
  \hline
CleanDish &
  4 &
  yes &
  2 &
  26.43s &
  150 &
  90\% &
  0\% \\
Gathering &
  3 &
  No &
  1 &
  12.93s &
  280 &
  78\% &
  0\%\\
CollectDish &
  3 &
  No &
  1 &
  24.79s &
  300 &
  85\% &
  0\% \\
\cline{1-8}
\end{tabular}

\caption{\textbf{Overview of task characteristics}: \textbf{Object Num} refers to the number of objects the robot interacts with; \textbf{Color} indicate whether the task involves classifying this feature; \textbf{Logic Step} represents the number of logical deductions required; \textbf{Avg Length} is the average duration of each demonstration video; \textbf{Demo Num} denotes the total number of demonstrations for each task.  
}
\label{tab:task_performance}
\end{table*}


To assess the influence of training frameworks and model architecture choices, we conducted an ablation study in which we systematically alternated between different training frameworks and models, and evaluated their performance across various tasks. The results, presented in Table \ref{tab:task_performance}, provide a comparative analysis of the effectiveness of Diffusion Policy (DP) \cite{c1} and Action Chunking with Transformers (ACT) \cite{c13}, as well as the performance of DP models trained using LeRobot versus CoinRobot. Our findings indicate that DP model trained within our framework outperform that trained with LeRobot. We also compare the performance of various DP model structures: ResNet18 with a CNN based module, and ResNet18, ResNet34, and FPN-based ResNet34 with transformer-based across different tasks.

Additionally, we observed that while ACT generally achieves high success rates on tasks with fixed item locations, it demonstrates limited generalizability when item locations vary, a capability in which DP models excel. These insights offer valuable directions for future research and hold practical implications for the development of robust robotic control systems.

\subsection{Model Generalization}
\label{sec:Model Generalization}
\subsubsection*{Multi-task Generalization}
\begin{figure}
    \centering
    \includegraphics[width=\linewidth]{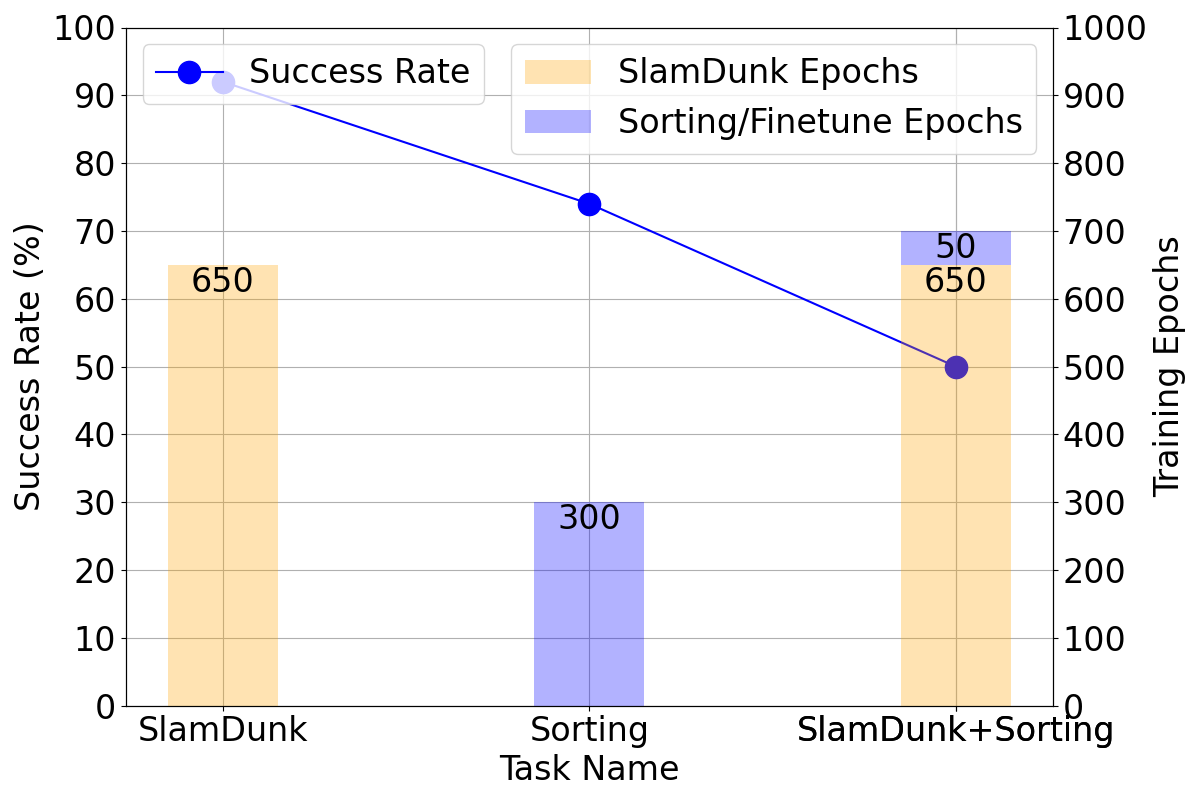}
    \caption{Comparison of training requirement and performance between individual task models and multitask model subsequently fine-tuned for only 50 epochs on a pretrained individual task model.}
    \label{fig:multitask generalization}
\end{figure}

Similar to the early stages of deep learning, prior work in imitation learning has predominantly focused on training models for single, specific tasks. In contrast, we propose a multi-task learning approach that leverages shared representations across tasks by fine-tuning a pre-trained model checkpoint on a combined dataset. Specifically, we initialize our model using a checkpoint trained on the ‘SlamDunk’ task for 650 epochs and subsequently fine-tune it on a dataset that combines both the ‘SlamDunk’ and ‘Sorting’ tasks. This process ensures that the model adapts to the new task while retaining knowledge from the original task, as the fine-tuning procedure optimizes the model weights using data from both tasks simultaneously. By doing so, the model achieves a balance between task-specific adaptation and generalization across tasks. 
Remarkably, the ‘Sorting’ task was successfully completed after only 50 additional epochs of fine-tuning, underscoring the efficiency and effectiveness of our multi-task learning strategy, as shown in Figure \ref{fig:multitask generalization}. It is important to note that the two tasks differ significantly in their setups. Specifically, the ‘SlamDunk’ task involves manipulating a toy ball and basket, while the ‘Sorting’ task requires handling blocks and plates. Additionally, the end-effector (eef) motion profiles differ between the tasks, particularly in terms of the height and target range required to complete the ‘SlamDunk’ task. These differences highlight the challenges of generalizing across tasks with distinct object geometries and motion constraints, further emphasizing the robustness of our approach.

\subsubsection*{Environmental Scene Generalization}
\begin{figure*}[t]
\centering

\begin{minipage}[b]{0.32\textwidth}
    \centering
    \includegraphics[width=\textwidth]{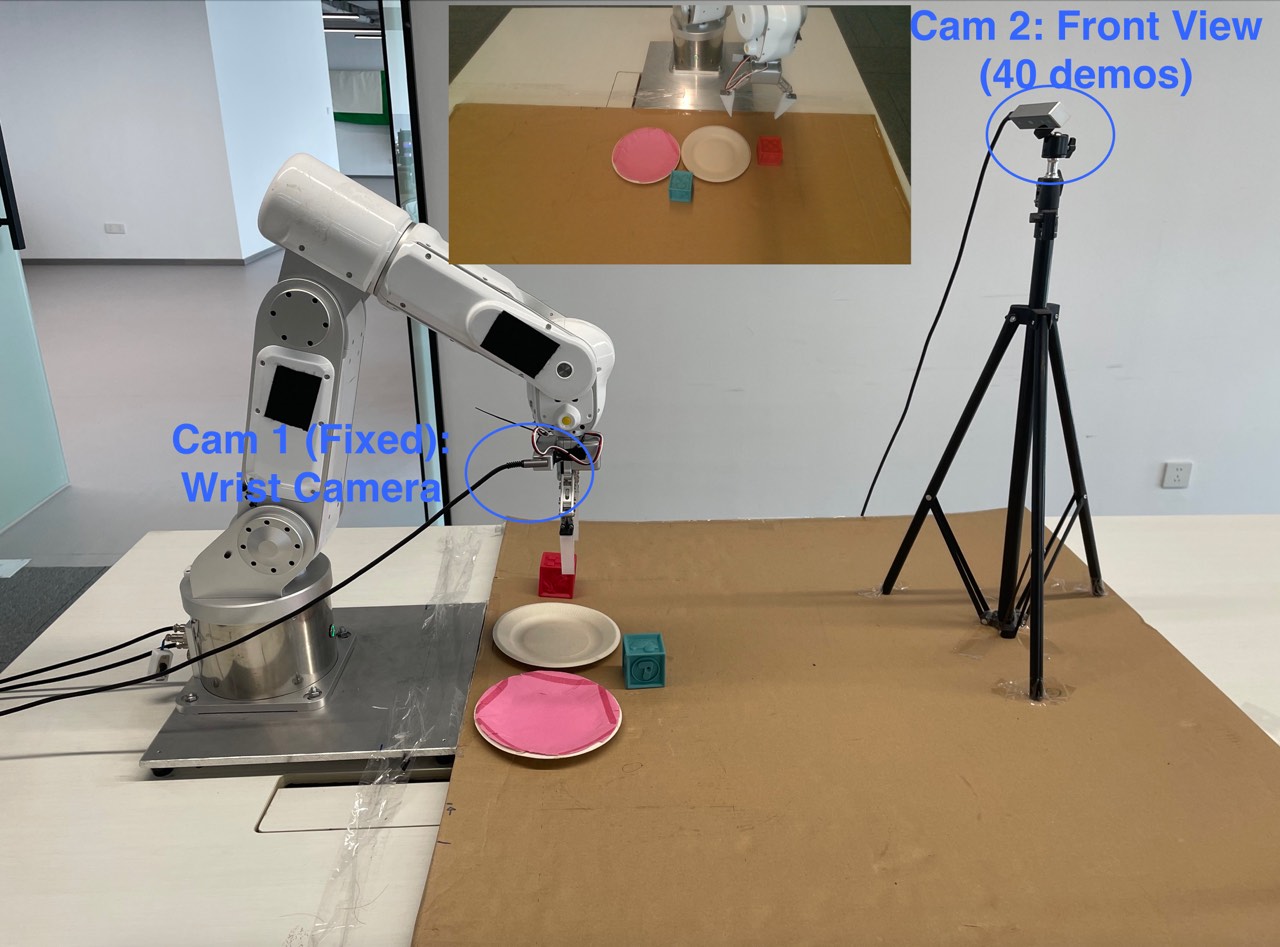}
    \caption*{(a) 2nd camera: front view 
(40 demos)}
\end{minipage}
\hfill
\begin{minipage}[b]{0.32\textwidth}
    \centering
    \includegraphics[width=\textwidth]{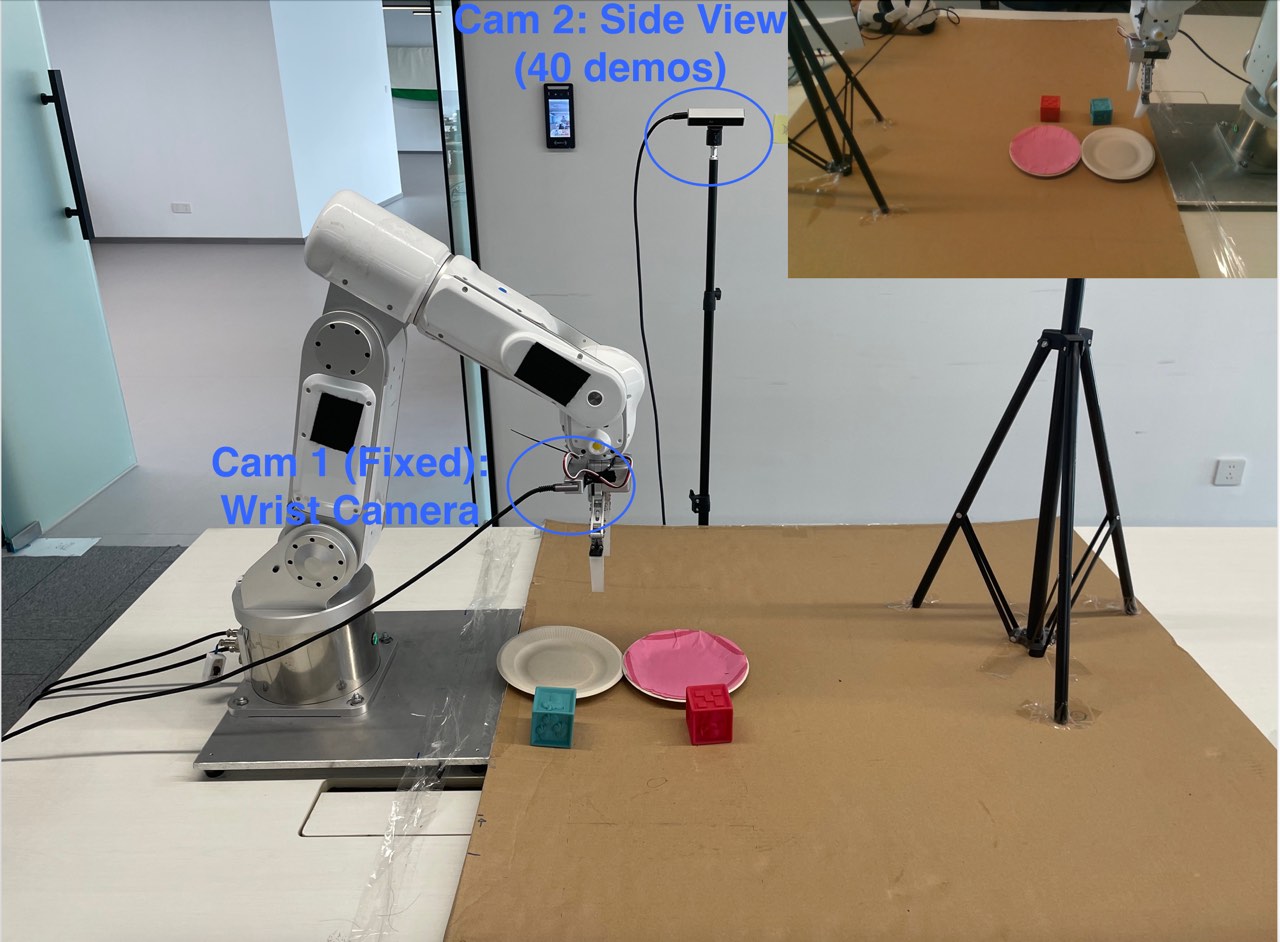}
    \caption*{(b) 2nd camera: side view (40 demos)}
\end{minipage}
\hfill
\begin{minipage}[b]{0.32\textwidth}
    \centering
    \includegraphics[width=\textwidth]{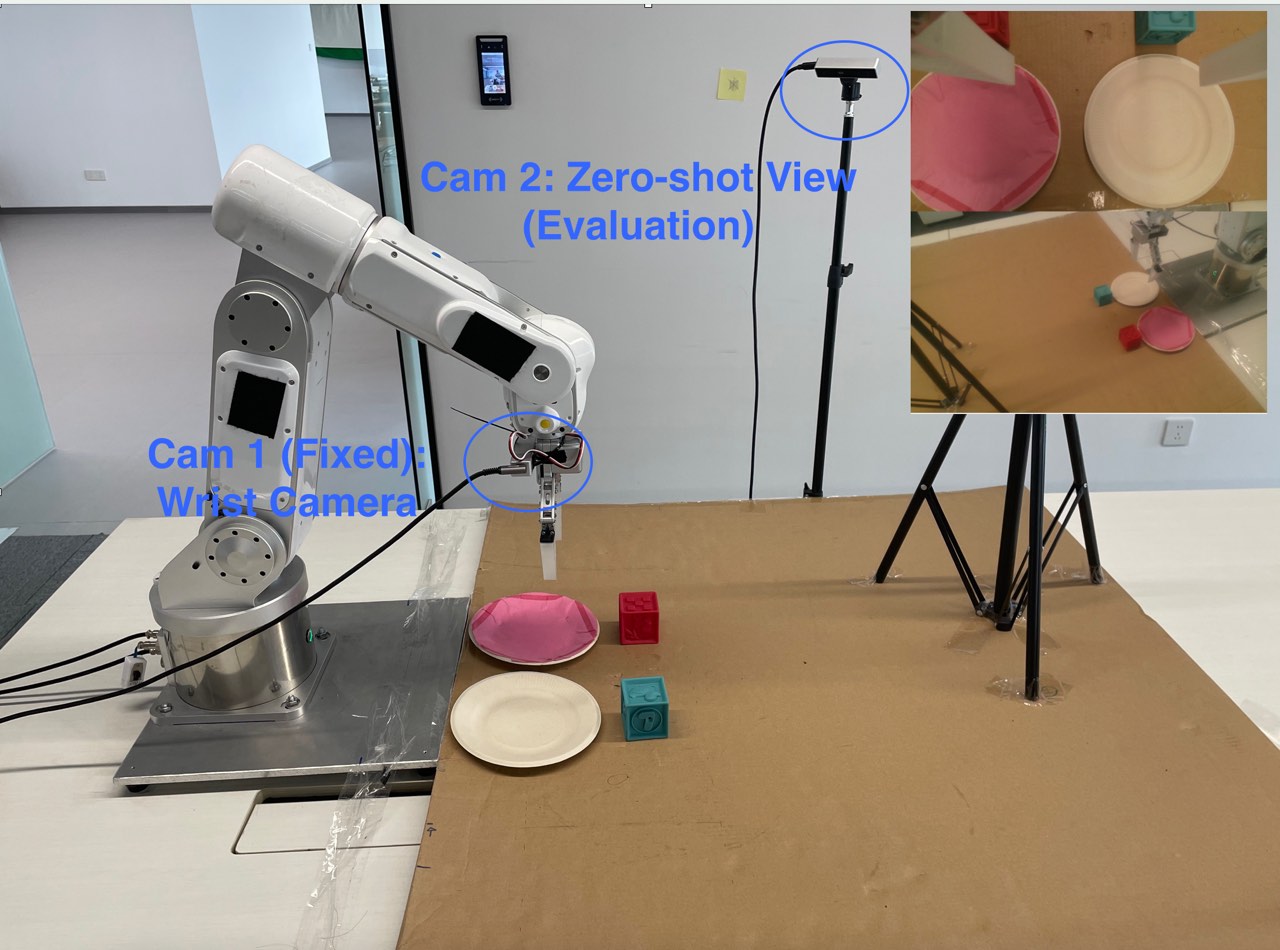}
    \caption*{(c) 2nd camera: zero-shot view (evaluation)}
\end{minipage}

\caption{Multi-view generalization performance. To evaluate robustness to novel camera perspectives, a fine-tuning phase was conducted using multi-angle visual data. The wrist camera remained fixed, while the second camera collected demonstration datasets from two distinct viewpoints: front (Position 1, 40 demonstrations) and side (Position 2, 40 demonstrations). The model was fine-tuned for an additional 100 epochs on this combined dataset. During evaluation, the fine-tuned model demonstrated successful generalization to zero-shot \cite{c53} camera configurations, including previously unseen positions and angles, in the 'Sorting' task. In contrast, the original fixed-view model failed under these novel conditions.
(a) \emph{Front-view training data}: Position 1 (40 demonstrations). (b) \emph{Side-view training data}: Position 2 (40 demonstrations). (c) \emph{Zero-shot evaluation}: The dual-view fine-tuned model is able to complete the task under unseen camera configurations (20\% success rate), whereas the fixed-view baseline model cannot (0\% success rate).}
\label{fig:multiview_tasks}
\end{figure*}

In standard imitation learning setups, demonstration data is typically collected with fixed camera positions, commonly involving one global view and another camera mounted on the end effector. This setup requires identical camera positions and view angles during inference to replicate the training conditions. Consequently, models trained under these constraints often exhibit limited adaptability and generalization to even minor variations in camera placement in real-world scenarios.
With one camera fixed at end-effector (eef), by incorporating merely two distinct camera views for the other (front/side view each providing 40 demonstrations), we fine-tuned the model for an additional 100 epochs. Notably, in the 'Sorting' task, this fine-tuning with multi-angle camera data enabled the model to complete task despite variations in camera positioning, including zero-shot \cite{c53} positions and angles that were entirely unseen during the training phase, as shown in Figure \ref{fig:multiview_tasks}. This result indicates that the model achieved certain level of generalization across a range of environmental visual conditions.


\section{Limitations and Future work}
\label{sec:Discussion and Future work}
Robotic learning should exhibit adaptability across diverse robotic platforms and industrial applications, avoiding reliance on overly complex or impractical configurations. While our framework provides a seamless pipeline from data collection to model deployment, facilitating efficient replication within a data-driven paradigm, a significant challenge remains the substantial data requirements for training. A key objective is to reduce the volume of data needed while maintaining robust model performance. One promising approach is to leverage pretrained models on large-scale, open-source robotic trajectory datasets, such as X-Embodiment \cite{c47}, which enable policies to adapt to novel tasks and environments with minimal fine-tuning. This strategy not only reduces data dependency but also enhances generalization across diverse scenarios.  In this work, we adopted a Diffusion Policy-inspired architecture, incorporating modifications to both the perception and action prediction modules. A promising direction for future research involves integrating human language instructions \cite{c43, c45, c46} to enhance the framework's logical reasoning capabilities, enabling more intuitive and context-aware task execution.

In summary, our future efforts will focus on progressively minimizing the data requirements of our framework by harnessing advanced transfer learning techniques and pretrained models. We anticipate that this approach will make a meaningful contribution to the robotics learning community, paving the way for more scalable and accessible solutions in the field of autonomous systems.

\section{Conclusion}

In this study, we present a comprehensive framework for deploying robotic systems, integrating task design, data collection, model training, and inference into a unified pipeline. Our platform can quickly adapt to new robotic hardware, data collection tool and model architecture, thereby broadening accessibility to cutting-edge research.

Our work also provides a systematic set of guidelines for task design, model evaluation metrics. Through extensive experimentation, we validated the feasibility of training multi-task models on real-world tasks and observed that even incremental modifications to model architectures can yield significant performance gains across a variety of tasks. These insights offer practical contributions to the optimization of model architectures for deployment in diverse and complex environments.

In conclusion, we propose a generalized robotic learning framework, supported by a dataset encompassing 7 real-world tasks, designed to advance the field of embodied intelligence. By promoting open-source collaboration and fostering research, our framework aims to catalyze the development of emergent capabilities in robotics, akin to the transformative impact of large-scale language models. We believe this work lays a foundation for future breakthroughs in autonomous systems and democratizes access to state-of-the-art robotic learning methodologies.




\bibliographystyle{unsrt}
\bibliography{Reference}

\end{document}